\documentclass[11pt,a4paper]{article}
\usepackage[hyperref]{emnlp2020}
\usepackage{times}
\usepackage{latexsym}

\usepackage{microtype}

\usepackage{amsfonts}
\usepackage{pgfplots}
\usepackage{graphicx}

\aclfinalcopy 

\title{Cross-lingual Word Embeddings beyond Zero-shot Machine Translation}

\author{Shifei Chen\\
 Dep. of Linguistics and Philology \\
 Uppsala University \\
 \texttt{\small{shifei.chen.2701@student.uu.se}} \\\And
 Ali Basirat \\
 Dep. of Linguistics and Philology \\
 Uppsala University \\
 \texttt{\small{ali.basirat@lingfil.uu.se}} \\}

\date{}

\begin{document}
\maketitle

\begin{abstract}
We explore the transferability of a multilingual neural machine translation model to unseen languages when the transfer is grounded solely on the cross-lingual word embeddings. Our experimental results show that the translation knowledge can transfer weakly to other languages and that the degree of transferability depends on the languages' relatedness. We also discuss the limiting aspects of the multilingual architectures that cause the weak translation transfer and suggest how to mitigate the limitations. 
\end{abstract}

\section{Introduction}
The multilingual neural machine translation (NMT) aims at training a single translation model between multiple languages \citep{Johnson:2016aa,aharoni-etal-2019-massively}.
Among the appealing points of multilingual NMT models are their ability for zero-shot learning, to generalize and transfer a translation model to unseen language pairs \citep{Johnson:2016aa}.
In zero-shot learning, a multilingual model trained on a set of language pairs is tested on an unseen language pair whose elements are still in the set of individual training languages.
The knowledge in this setting is transferred across the shared parameters of the model. 

\citet{Kim:2019aa} argues that one of the critical components responsible for the knowledge transfer in multilingual NMT is the embedding layers which train a cross-lingual vector space for all words of the languages.
They show that multilingual translation models can be transferred to new languages if the model's cross-lingual vector space is aligned to the new language's vector space.
However, the success of their approach is at the cost of aligning the word vectors and retraining the new language's translation model.

This research studies the importance of word representation in the multilingual NMT transfer model in a more controlled setting based on the pre-trained cross-lingual word embeddings \citep{Bojanowski:2016aa,Ammar:2016aa,Joulin:2018aa,Ruder:2019aa}.
More specifically, we examine the transferability of a multilingual NMT when it is applied to a new test language. 
We use cross-lingual word embeddings as the source of transfer knowledge to the test languages and leave the translation model's shared parameters to model the interrelationships between the training languages.
Our setting is different from the zero-shot setting of \citet{Johnson:2016aa} in the way that one of the test languages does not belong to the set of individual training languages.
It is also different from the \citet{Kim:2019aa}'s setting, in that it does not retrain the embeddings and model parameters. 

We hypothesize that a multilingual NMT model trained with pre-trained cross-lingual word embeddings should transfer reasonably to a new test language if the word representation has any role in the model's transferability.
Our preliminary results on a set of test languages show that the translation model transfers only weakly to the unseen languages, and the amount of the transferability depends on the similarity of the test language to the training languages.

\section{Related Work}
The transferability of multilingual neural machine translation models is vital from both the theoretical and practical perspectives.
The theoretical importance of these models come to the way that they find the correspondence between language pairs \citep{Johnson:2016aa,lu-etal-2018-neural}.
The practical importance is due to their effective use for the translation of low-resource languages \citep{Zoph:2016aa,Nguyen:2017aa}.
\citet{Johnson:2016aa} shows that multilingual NMT trained on a massive training set can generalize reasonably well to the zero-shot learning setting.
This capability is further examined by \citet{aharoni-etal-2019-massively}, demonstrating that multilingual NMT models transfer better when trained on a massive training set.
\citet{Kim:2019aa} shows that multilingual NMT models can be transferred to a new language when their embedding spaces are realigned to the embeddings of the new language.

Cross-lingual word embeddings are important for the translation of low-resource languages.
They could either be used as external auxiliary information \cite{Conneau:2017aa, Lakew:2019aa}, or as the embedding layer directly \cite{neishi-etal-2017-bag,artetxe-etal-2017-learning}.
\citet{Qi:2018aa} explored how effective are aligned pre-trained word embeddings in an NMT system.
They found that regardless of languages, alignment is useful as long as it is applied in a multilingual setting.
They believe that since both the source and the target side vector spaces are already aligned, the NMT system will learn how to transform a similar fashion from the source language to the target language.
It is then interesting to see how far can aligned word embeddings could go beyond known languages.
Zero-shot translation analyzed this question by testing the multilingual NMT system on unseen language pairs --- language in either source or target side of the translation is known to the system, but their paired combination remains unknown.
In this work, we would like to take a step further to see how aligned word embeddings would work for languages that are entirely unseen to the multilingual NMT setting.

\section{Multilingual NMT}
Multilingual Neural Machine Translation (NMT) \cite{Johnson:2016aa} extends the attentive encoder-decoder framework of NMT \citep{Bahdanau:2014aa} to translate between multiple languages simultaneously. 
Instead of training multiple bilingual NMT models, the multilingual NMT augments the input representations with a language indicator at the beginning of each training sentence, to be trained in an end-to-end fashion. 
Despite its simplicity, the multilingual NMT has shown significant improvement to the machine translation \citep{aharoni-etal-2019-massively} and provides for zero-shot translation -- translating between an unseen language pair during the training time, which benefits low-resource languages by transferring knowledge from their high-resource relatives \citep{Zoph:2016aa,Nguyen:2017aa}.

\section{Cross-lingual Word Embeddings}
Cross-lingual word embeddings represent lexicons from several different languages in a shared embedding space, providing for the word-level language transfer models and the cross-lingual study of words. 
\citet{Ruder:2019aa} collects a survey of different approaches used to train cross-lingual word embeddings.
One of the main approaches is to find a mapping between monolingual embeddings spaces based on a seed dictionary. The fastText cross-lingual embeddings are among the extensively used resources trained in this way, using the monolingual word embeddings of \citet{Bojanowski:2016aa} and the mapping approach of \citet{Joulin:2018aa}. 

\section{Experiments}
We study the transferability of multilingual NMT based on two disjoint sets of languages to train and test a translation model.
We interpret the average BLEU score \cite{papineni-etal-2002-bleu}, and individual accuracy scores on 1-3 grams of both translation directions from the test languages, as the transferability measure of the translation model to the unseen test languages. That is to say; we consider high values of BLEU score and individual 1-3 gram accuracy as the goodness of the model transfer from the training languages to the test languages. The test languages remain unseen to the translation model during the training phase. 

We choose English (EN), German (DE), and French (FR) as the training languages together with Swedish (SV), Hungarian (HU), and Hebrew (HE) as the test languages. The languages are selected to analyze the effect of language similarity to the model transferability. Among the test languages, Swedish is the most similar one to the training languages. Hence, it is expected to obtain a higher BLEU score on Swedish than the two other test languages if language similarity plays any role in the models' transferability. On the other hand, Hebrew should get the lowest score since it is more different from the training languages. 


\begin{table}
  \centering
  \begin{tabular}{r|rrr}
    \hline
    Language & train & dev & test \\ [0.25ex]
    \hline\hline
    EN+DE+FR & 1013478 & $-$ & $-$ \\
    +SV & $-$ & 9390 & 12423 \\
    +HU & $-$ & 20332 & 25606 \\
    +HE & $-$ & 24554 & 28546 \\
    \hline
    EN+DE+DA & 491537 & $-$ & $-$ \\
    +SV & $-$ & 8037 & 9344 \\
    \hline
    +NL & 1225511 & $-$ & $-$ \\
    +NL+SV & $-$ & 11126 & 13378 \\
    \hline
    +NL+NO & 1322133 & $-$ & $-$ \\
    +NL+NO+SV & $-$ & 12304 & 14430 \\
    \hline
  \end{tabular}
  \caption{Number of sentences in each language combination after preprocessing}
  \label{table:corpus_size}
\end{table}
We used the TED talk subtitle corpus \citep{Qi:2018aa} to train, validate, and test the multilingual NMT model.\footnote{\url{https://github.com/neulab/word-embeddings-for-nmt}} 
We downcase all letters in the corpora and removed sentences longer than $60$ words as well as less frequent words that appeared only once.
Table \ref{table:corpus_size} shows the corpus size of each language combinations after prepossessing. Each translation model is trained on the training data. The development data is used for early stopping the training procedure and the the test data is used to measure the transfer quality on the test languages.\footnote{We have not seen a significant difference between using and not using the development set for early stopping.} 
Our neural network is a modified version of the one from \citet{Qi:2018aa}, which was built upon XNMT \cite{Neubig:2018aa}. The only change we have made is doubling the encoding layer to a 2-layer-bidirectional LSTM network in order to accommodate the additional information in the multilingual scenario. Everything else is the same as the original experiment settings, including the encoder-decoder model with attention \cite{Bahdanau:2014aa}, beam search size of $5$, batches of size $32$, dropout at $0.1$, the Adam optimizer \cite{Kingma:2014aa}. The initial learning starts at $0.0002$ and decays by $0.5$ when the BLEU score on the development set decreases \cite{Denkowski:2017aa}.

The fastText word embeddings are used as the source of the knowledge transfer across languages.
\footnote{\url{https://fasttext.cc/docs/en/aligned-vectors.html}} 
We concatenated the embeddings of all languages in both the training and test sets into a single file, and we used it to initialize the embedding layers of the NMT model.
The embedding layers of the networks are kept frozen during the training to preserve the training embeddings in their original embeddings space, the same space as the test embeddings.  
We use all embeddings with no change except those related to the words with the same form in multiple languages.
The vector of these words is set to the mean vector of the individual word vectors in each language.  


\section{Preliminary Results}

\begin{table}
 \centering
 \begin{tabular}{r|*{4}{l}}
 \hline
 Language & BLEU & 1gram & 2gram & 3gram \\ [0.25ex]
 \hline\hline
 EN+DE+FR & 29.2 & 0.57 & 0.34 & 0.24 \\ 
 \hline
 SV & 1.5 & 0.16 & 0.02 & 0.00 \\ 
 HU & 1.1 & 0.17 & 0.02 & 0.00 \\
 HE & 1.0 & 0.16 & 0.02 & 0.00 \\
 \hline
 \end{tabular}
 \caption{The translation performance on the training languages (EN+DE+FR), and and each of the test languages.}
 \label{table:initial_results}
\end{table}

Table \ref{table:initial_results} summarizes the translation performance on the test division of the training languages altogether (EN+DE+FR) and each of the test languages (SV, HU, and HE). 
The relatively low results on the test languages compared with the training languages indicate that cross-lingual embeddings are not rich enough for the model transfer in machine translation.
However, when it comes to a random setting with no pre-trained embeddings, we see that the translation model trained with cross-lingual embeddings performs substantially better (Avg BLEU=1.2) than a model trained with random embeddings (BLEU=0.1).


The slightly better result on Swedish suggests that the model transfers better to similar languages. 
We continue our experiment with more languages to further study the effect of language similarity. 
For this purpose, we homogenize the training languages more toward Swedish by excluding French and adding Danish (DA), Dutch (NL), and Norwegian (NO) one-by-one to the training set. 

\begin{table}
 \centering
 \begin{tabular}{r|llll}
 \hline
 Language & BLEU & 1gram & 2gram & 3gram \\ [0.25ex]
 \hline\hline
 EN+DE+FR & 1.5 & 0.16 & 0.02 & 0.00 \\
 \hline
 EN+DE+DA & 4.1 & 0.28 & 0.07 & 0.02 \\
 +NL & 3.3 & 0.23 & 0.05 & 0.02 \\
 +NL+NO & 4.7 & 0.31 & 0.07 & 0.03 \\
 \hline
 \end{tabular}
 \caption{The transfer results to Swedish.} 
 \label{table:language_similarity}
\end{table}

Table~\ref{table:language_similarity} summarizes the transfer results from each of the training settings to Swedish. 
We see that the homogeneous setting performs better than the other setting that includes French in the training set. 
More specifically, when adding DA nad NO to the training language, both the BLEU score and individual accuracy scores are improved, while adding NL to the training language worsened the results. We speculate that the large shared vocabulary between Swedish, Danish, and Norwegian drives the performance increase. 
When we distinguish each word with its language origin, the result dropped to 1.7 BLEU score again (for the EN+DE+DA experiment, tested on language SV).
This observation further strengthens the effect of the shared vocabulary on transfer learning. 

\subsection{Discussion}
Our preliminary results on the transferability of multilingual NMT models to unseen languages show that these models can transfer weakly to completely unseen languages if the transfer learning is grounded on the cross-lingual word embeddings. 
One reason for the weak translation transfer across languages is that the output space of the model's decoder is not aligned to the cross-lingual embeddings' space. This is because of the many transformations applied to the input vectors during the translation process. It can be seen if we compare the BLUE score for each side of the translations, say SV$\to$EN+DE+DA versus EN+DE+DA$\to$SV. We get a relatively higher BLEU score of 6.0, when translating from the unseen language (SV$\to$EN+DE+DA), but almost no translation (BLUE=1.0) is performed on the opposite direction when translating to the test language (EN+DE+DA$\to$SV). 

Furthermore, a large amount of error in the output is because the output layer of the model's decoder does not provide any mechanism to transfer the translation knowledge between languages. The layer has one entry per each word in the entire vocabulary set. The entries are activated independently, one at each time. Since the model does not see any example from the unseen test languages during the training phase, the output connections corresponding to the unseen words are down-weighted during the training phase. Hence, it is improbable for the model to output words from the unseen languages, except for those shared with the training languages. 

The above discussion suggests that the multilingual NMT architecture of \citet{Johnson:2016aa} might transfer better to unseen languages if the decoder and the encoder embeddings are in the same vector spaces. A simple way to reach such an alignment between the two embedding spaces is to add a regularization cost based on the divergence of the two spaces from each other to the loss function of the multilingual model. We consider this as a future step for this research. Another potential step to be explored in the future is to provide the translation model with information about the language relatedness, either to use language embeddings \citep{littell-etal-2017-uriel} together with the cross-lingual embeddings, or to constrain the model's output to the desired target language by re-aligning the output vector space back to the input vector space. Moreover, a more in-depth error analysis is required to address the other potential limitations of the multilingual model transfer based on the cross-lingual embeddings.

Finally, we would like to emphasize that this is still an ongoing research and some caveats about the results. We examine the translation transfer on a relatively small set of languages with a more in-depth analysis of only one language. It will be interesting to consider a more extensive language set and study how the model transfer perform in different language families. We have tested only one set of cross-lingual embeddings on an attention-based encoder-decoder NMT architecture. Although we believe it is unlikely to see marginally different results from other sets of embeddings (especially when it comes to conventional word vectors), it is still worth exploring how other sets of embeddings (e.g., the multilingual contextualized cross-lingual embeddings \cite{devlin-etal-2019-bert} and the multilingual sub-word embeddings \cite{heinzerling2018bpemb} perform in this scenario. It will also be interesting too see if other NMT architectures, i.e., Transformers \cite{Vaswani:2017aa}, would bring any improvements.

\bibliographystyle{acl_natbib}
\bibliography{thesis}

\end{document}